\documentclass[letterpaper]{article} 
\usepackage{aaai23}  
\usepackage{times}  
\usepackage{helvet}  
\usepackage{courier}  
\usepackage[hyphens]{url}  
\usepackage{graphicx} 
\usepackage{natbib}  
\usepackage{caption} 
\usepackage{algorithm}
\usepackage{algorithmic}
\usepackage{booktabs}
\usepackage{bm}
\usepackage{amssymb}
\usepackage{amsmath}
\usepackage{color}

\usepackage{newfloat}
\usepackage{listings}
\DeclareCaptionStyle{ruled}{labelfont=normalfont,labelsep=colon,strut=off} 
\floatstyle{ruled}
\newfloat{listing}{tb}{lst}{}
\floatname{listing}{Listing}
\title{Robust One-shot Segmentation of Brain Tissues via Image-aligned Style Transformation}
\author{
    Jinxin Lv\textsuperscript{\rm 1,2 \thanks{Co-first authors contributing equally. $^{\dag}$ Corresponding authors.}},%
    Xiaoyu Zeng\textsuperscript{\rm 1,2 \footnotemark[1]},
    Sheng Wang\textsuperscript{\rm 1,2},
    Ran Duan\textsuperscript{\rm 1,2},
    Zhiwei Wang\textsuperscript{\rm 1,2 \footnotemark[2]},
    and Qiang Li\textsuperscript{\rm 1,2 \footnotemark[2]} 
}

\affiliations{
    \textsuperscript{\rm 1} Britton Chance Center for Biomedical Photonics, Wuhan National Laboratory for Optoelectronics, Huazhong University of Science and Technology, Wuhan, China\\
    \textsuperscript{\rm 2} MoE Key Laboratory for Biomedical Photonics, Collaborative Innovation Center for Biomedical Engineering, School of Engineering Sciences, Huazhong University of Science and Technology, Wuhan, China\\
    \{jinx, zengxiaoyu, wagnsheng, duanran1998, zwwang, liqiang8\}@hust.edu.cn
}

\usepackage{bibentry}

\begin{document}

\maketitle

\begin{abstract}
One-shot segmentation of brain tissues is typically a dual-model iterative learning: a registration model (reg-model) warps a carefully-labeled atlas onto unlabeled images to initialize their pseudo masks for training a segmentation model (seg-model); the seg-model revises the pseudo masks to enhance the reg-model for a better warping in the next iteration. However, there is a key weakness in such dual-model iteration that \emph{the spatial misalignment} inevitably caused by the reg-model could misguide the seg-model, which makes it converge on an inferior segmentation performance eventually. In this paper, we propose a novel \emph{image-aligned style transformation} to reinforce the dual-model iterative learning for robust one-shot segmentation of brain tissues. Specifically, we first utilize the reg-model to warp the atlas onto an unlabeled image, and then employ the Fourier-based amplitude exchange with perturbation to transplant the style of the unlabeled image into the aligned atlas. This allows the subsequent seg-model to learn on the aligned and style-transferred copies of the atlas instead of unlabeled images, which naturally guarantees the correct spatial correspondence of an image-mask training pair, without sacrificing the diversity of intensity patterns carried by the unlabeled images. Furthermore, we introduce a feature-aware content consistency in addition to the image-level similarity to constrain the reg-model for a promising initialization, which avoids the collapse of image-aligned style transformation in the first iteration. Experimental results on two public datasets demonstrate 1) a competitive segmentation performance of our method compared to the fully-supervised method, and 2) a superior performance over other state-of-the-art with an increase of average Dice by up to 4.67\%. The source code is available at: \url{https://github.com/JinxLv/One-shot-segmentation-via-IST}.

\end{abstract}

\section{Introduction}

Accurate segmentation of brain magnetic resonance imaging (MRI) images is a fundamental technique for measuring the volume of brain tissues, and assisting neurosurgeons in analyzing, judging, and treating diseases \cite{geuze2005mr}. The existing fully supervised segmentation methods can achieve promising segmentation accuracy only when plentiful, high-quality labeled data is available \cite{zhao2019data}. However, the structure of brain tissues is extremely intricate, which makes the manual annotation of 3D brain MRI images time-consuming, error-prone and expertise-required. For example, an experienced neurosurgeon usually takes hours to annotate one brain MRI 3D image \cite{klein2012101}.

To alleviate the negative impact of the resulting scarce labels in brain tissue segmentation, numerous label-efficient solutions have been put forward. The most straightforward one is data augmentation to increase the number of data and labels from those on hand. For example, a group of methods \cite{christ2016automatic,zhang2019unseen,3dunet,vnet} utilized traditional strategies like affine transformation and global intensity transformation to enlarge training data quantity. Besides, there is another group of methods \cite{zhao2019data,chen2020realistic} proposed to introduce deep learning-based techniques to increase both data quantity and diversity. For example, Generative Adversarial Networks (GANs) are often utilized to `create' new training samples via image synthesis. However, these methods require an extra non-trivial computational overhead, which limits the flexibility and convenience of practical usage.

Another well-known solution is called atlas-based segmentation \cite{wang2021alternative,wang2020lt,lv2022joint,wang2019patch,coupe2011patch}, which enables a label-efficient segmentation with the help of medical image registration. The basic idea of these methods is to `spread' the segmentation label of an atlas (i.e., a special sample with high image quality for accurate manual annotation) onto a target image that is about to be segmented. The label spreading process is equivalent to warping the atlas to align with the target image by a registration model. Thus, the simultaneously warped label mask can be treated as the output segmentation mask. Since the registration model can be trained in an unsupervised manner by maximizing the image-level similarity, e.g., mutual information, between the warped atlas and target image, the atlas-based segmentation methods only require a few or even a single labeled atlas in the inference phase.

However, some recent methods \cite{xu2019deepatlas,li2019hybrid,he2020deep,estienne2020deep,estienne2019u,beljaards2020cross} pointed out that the atlas-based segmentation usually leads to sub-optimal results, since it learns indirect image-to-image warping relation. They instead proposed to train an additional segmentation model to directly learn the image-to-mask mapping relation, while the registration model is just used to initialize a set of pseudo masks of unlabeled training images via the label spreading process as mentioned above. Concretely, a dual-model iterative learning is employed. First, a registration model (reg-model) warps an elaborately-labeled atlas, and is trained by maximizing the image-level similarity between the warped atlas and the target image. Second, the trained reg-model `spreads' the label of the atlas onto the unlabeled images to initialize their pseudo masks, and thus the seg-model can be trained using the image-mask pairs. Third, the trained seg-model refines the initialized pseudo masks, and triggers the next training iteration, in which the reg-model can be enhanced under extra supervision to minimize the Dice loss derived from the refined pseudo masks. Such dual-model iteration is named one-shot segmentation in the related works, since only one manual label is necessary for learning the image-to-mask mapping relation.

However, the above-mentioned methods coincidentally ignored a main flaw in the one-shot segmentation of brain tissues, which makes the seg-model not always converge on a desired segmentation performance. That is, the assigned pseudo masks are not perfectly aligned with the target images because of inevitable registration errors. Therefore, the resulting image-mask spatial misalignment could mess up the learning of the seg-model, misguiding the following iterations consequently.

\begin{figure*}[!t]
\centerline{\includegraphics[width=0.9\textwidth]{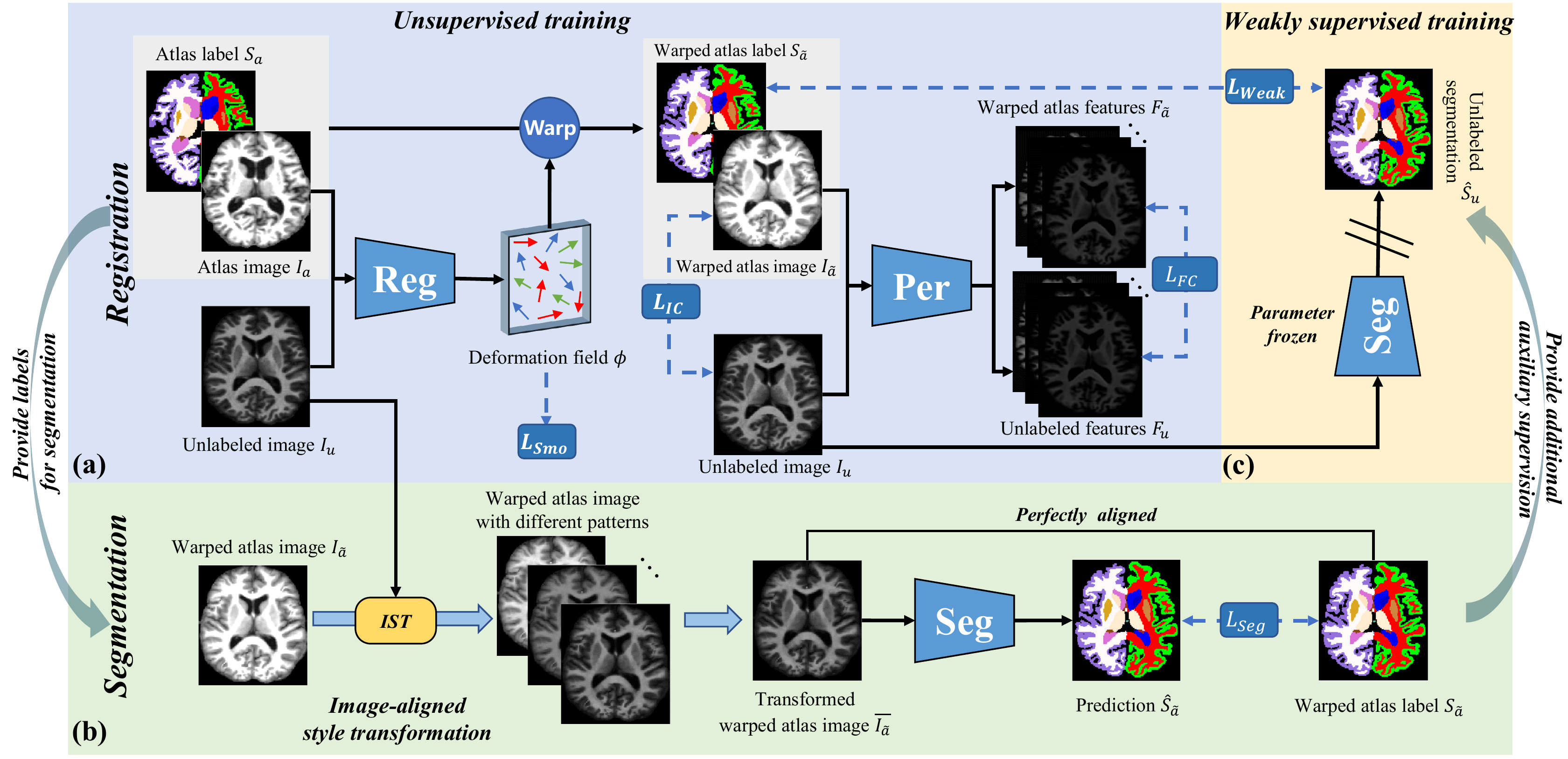}}
	\caption{The overview of the iterative and complementary one-shot segmentation framework. (a) The reg-model is first trained in an unsupervised manner, constraining the image and feature consistency to provide accurate initialization. (b) We then utilize the transformed warped atlas via image-aligned style transformation with its label to train the seg-model. (c) The revised prediction can be further exploited to weakly supervised the reg-model, triggering a new round of iteration.}
	\label{framework}
\end{figure*}

To address this, a simple strategy is to directly train the seg-model on the warped copies of atlas-mask pair, but this comes with a great sacrifice of pattern diversity. In this paper, we aim at a robust one-shot segmentation, and propose a novel  \emph{image-aligned style transformation}, which can well guarantee both correct spatial correspondence of image-mask pairs and a wide variety of image patterns. Specifically, we first utilize the reg-model to warp the atlas. The warped atlas and target image share the aligned brain structures (i.e., similar spectrum phase), but have different image patterns (i.e., different spectrum amplitude). We then employ the Fourier-based amplitude exchange to transform the style of the warped atlas to that of the target image by transplanting the target amplitude component into the warped atlas. During the amplitude exchange, a random factor is multiplied with the transplanted target component to further increase the variety of style-transformed copies of the atlas.

Nevertheless, the dissimilar image patterns between the atlas and unlabeled images have a certain possibility to  train an unreliable reg-model in the first iteration, which could make the style-transformed atlas corrupted due to the large registration errors. In view of this, we introduce an additional feature-aware content consistency to increase the reg-model robustness. Besides maximizing the low-level intensity consistency, we utilize a trained convolutional neural network (CNN) to extract semantic features from the warped atlas and the target image, and calculate the perceptual loss (i.e., the distance between the two feature maps) to constrain the high-level content consistency.

In summary, our contributions are listed as follows:
\begin{itemize}
    \item{We propose a novel \emph{image-aligned style transformation} (IST) to overcome the spatial misalignment, which plagues the current dual-model iterative methods of brain tissue one-shot segmentation, and to address the `either/ or' dilemma between the correct spatial correspondence and the diverse training patterns by inheriting the styles from unlabeled images.}
    
    \item{We additionally force a \emph{feature-aware content consistency} (FCC) in the training of reg-model to guarantee a foolproof initialization of pseudo masks, thus for a proper image-aligned style transformation in the first iteration. IST and FCC are facilely embedded into an iterative and complementary framework for robust one-shot brain tissue segmentation.}
    
    \item{Experimental results on two public datasets demonstrate that the dual-model iteration, which is equipped with image-aligned style transformation for seg-model and feature-aware content consistency for reg-model, achieves a superior performance of one-shot segmentation of brain tissues over other state-of-the-art methods.}
    
\end{itemize}

\section{Related Work}

\subsubsection{Atlas-based segmentation method} The atlas-based segmentation is essentially a registration task. It registers the labeled image (i.e. atlas) to the target image, and the labels are propagated to the target image to obtain the segmentation results. Recently, many learning-based registration methods have been proposed to boost the performance of atlas-based segmentation. For example, the methods \cite{vm,vm-diff} proposed a U-Net-like network named VoxelMorph to predict the deformation field, and combined image similarity and smoothness of the predicted field as losses for unsupervised training. The method \cite{wang2020lt} proposed LT-Net, utilizing the forward-backward consistency between the atlas and target images to stabilize the training process. However, these methods ignored that it is difficult to accurately align the given two images at once, especially in the brain image with complex tissue structure. To address this, some methods \cite{dual-MICCAI,wang2021alternative,dual-MIA,VTN,RCN,mok2020large,wang2021progressive,lv2022joint,hu2022recursive} proposed to decompose the target deformation field by multi-scale CNNs or multiple cascaded CNNs models. Despite achieving good registration performance, such atlas-based segmentation methods are susceptible to tissue gray-scale blurring, resulting in inaccurate segmentation results, since they only rely on the similarity between images and lack guidance of the anatomical structures. 

\subsubsection{One-shot segmentation in medical image} 
The one-shot segmentation, i.e., the joint segmentation with registration method, has been studied for decades based on the traditional approach, such as \cite{yezzi2001variational,pohl2006bayesian,mahapatra2010joint}. In recent years, many algorithms have been proposed to exploit deep CNNs to achieve one-shot segmentation, solving the time-consuming problem of traditional methods. The method \cite{xu2019deepatlas} first generated pseudo masks for unlabeled images based on the registration network and then used the generated image-mask pair to train the segmentation network. These two networks were mutually improved through iterative training. The method \cite{beljaards2020cross} proposed a multi-task network to improve the performance of two sub-networks, i.e., one for registration and another for segmentation, through cross-stitch feature sharing. The method \cite{he2020deep} proposed a joint training network based on registration and segmentation called deepRS, which utilized the discriminator \cite{gan,patchgan} to suppress the unaligned regions caused by the registration errors, thus boosting the performance of segmentation. Despite their success, the issue of spatial misalignment of image-mask pair is not eradicated. Moreover, it also brings additional computational overhead and the training process is more cumbersome.


\section{Method}

As shown in Figure \ref{framework}, our iterative and complementary framework for one-shot brain tissues segmentation consists of three parts: (a) registration with additional feature-aware content consistency, (b) segmentation with transferred image-mask pair via image-aligned style transformation, and (c) weakly supervised registration for triggering a new round of iteration. Below we introduce these three parts and the implementation details.


\begin{figure*}[!t]
	\centerline{\includegraphics[width=0.93\textwidth]{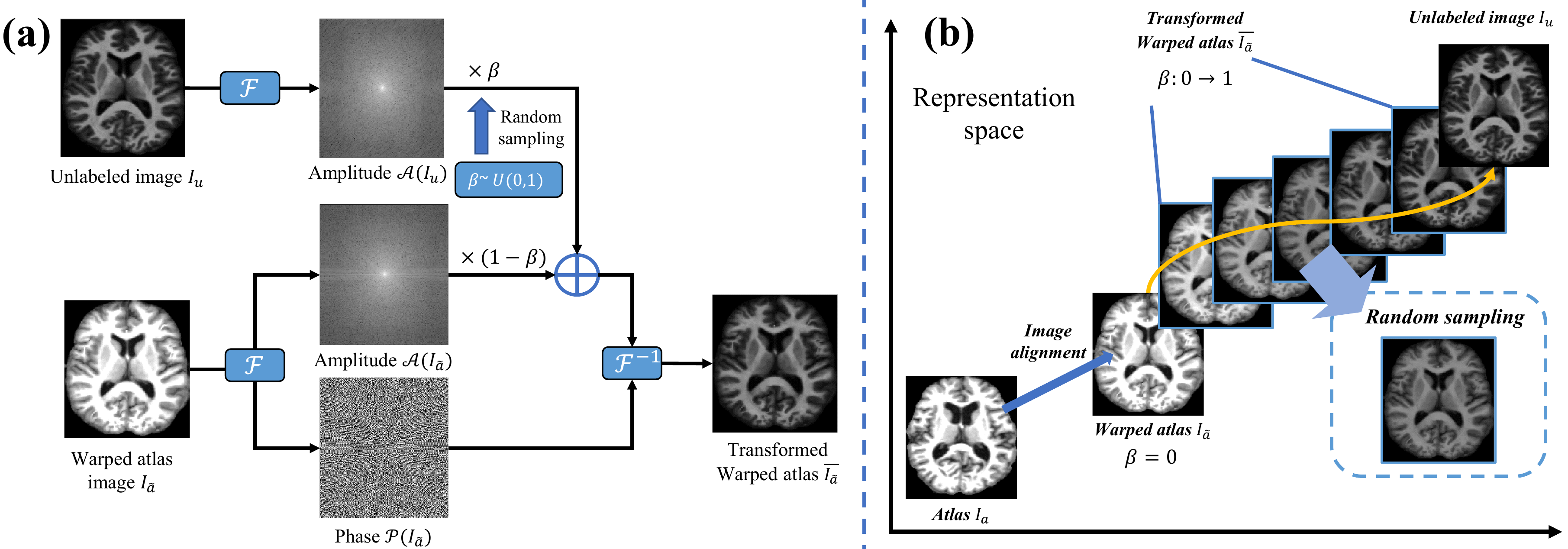}}
	\caption{(a) The detailed implementation of our proposed image-aligned style transformation. We first obtain two roughly aligned images, i.e., $I_{\tilde{a}}$ and $I_u$, and then the Fourier-based amplitude exchange with a perturbation of $\beta  \in [0,1]$ is employed to change the style of the $I_{\tilde{a}}$ to that of $I_{u}$. (b) Illustration of atlas transformed by image-aligned style transformation module in the representation space.}
	\label{fourier}
\end{figure*}

\subsection{Feature-aware content consistency in registration for better initialization }
Given a few pairs of atlas $I_a$ and unlabeled image $I_u$, we utilize the CNNs to learn the mapping from these image pairs to the deformation field $\phi$. The atlas is warped by the deformation field to guarantee the corresponding anatomical points between the warped atlas $I_{a} \circ \phi$ and  unlabeled image $I_u$ to be uniformly aligned. We denote the $I_{a} \circ \phi$ as $I_{\tilde{a}}$ for convenience.

Typically, the initial reg-model is trained by optimizing the image similarity loss between the warped atlas and the unlabeled image. However, the registration may fail due to the different patterns between the atlas-unlabeled image pair, resulting in the collapse of the subsequent image-aligned style transformation in the first iteration. In view of this, we trained the reg-model through a hybrid loss function, which combines the similarity loss of image-wise and feature-wise to guarantee the consistency of image and feature-aware content, aiming at improving the registration accuracy and achieving good initialization results for the subsequent training. 

Specifically, as shown in Figure \ref{framework}(a), given a deformation field $\phi$ predicted by reg-model, we first calculate image-wise similarity loss between the warped atlas $I_{\tilde{a}}$ and unlabeled image $I_u$ to ensure their image consistency, which is calculated as: $\mathcal{L}_{IC} = - NLCC\left( {I_{\tilde{a}}}, I_{u} \right)$. The $NLCC(\cdot, \cdot)$ represents the Normalized Local Correlation Coefficient, and the NLCC of the two images can be formulated as:

\begin{equation}\label{NLCC}
\resizebox{0.85\columnwidth}{!}{
$\begin{aligned}
    &NLCC\left( {X,Y} \right) =  \\
    &{\sum\limits_{p \in \Omega}\frac{\left( {\sum\limits_{p_{i} \in p}{( {X\left( p_{i} \right) - \overline{X_{l}}(p)} )( {Y\left( p_{i} \right) - \overline{Y_{l}}(p)} )}} \right)^{2}}{\left( {\sum_{p_{i} \in p}( {X\left( p_{i} \right) - \overline{X_{l}}(p)} )^{2}} \right)\left( {\sum_{p_{i} \in p}( {Y\left( p_{i} \right) - \overline{Y_{l}}(p)})^{2}} \right)}},
\end{aligned}$
}
\end{equation}
where $ \overline{X_{l}}(p)$ and  $ \overline{Y_{l}}(p)$ denote images with local mean intensities: $\overline{X_{l}}(p)= \frac{1}{n^{3}}{\sum_{p_{i} \in p}{X\left( p_{i} \right)}} $, the $p_{i}$ iterates over a $n^3$ volume around the image patch $p$, with n = 8 in  our experiments.

Besides, we utilized a perceptual CNN, which is a pre-trained model with the same architecture as our reg-model, to extract the semantic features with original image dimensions from the warped atlas and unlabeled images. These pattern-independent features are denoted as $F_{\tilde{a}}$ and $F_{u}$ respectively, and are used to guarantee feature-aware content consistency, denoting as: $\mathcal{L}_{FC} = - NLCC\left( F_{\tilde{a}},F_{u} \right)$.


Thus, the total loss function of the initial reg-model is formulated as:
\begin{equation}\label{FCC}
	\mathcal{L}_{Reg}^{0} = \mathcal{L}_{IC} + \mathcal{L}_{FC} + \lambda \mathcal{L}_{Smo},
\end{equation} 
where $\mathcal{L}_{Smo}=\sum\limits_{p_{i} \in \Omega}\left\| {\nabla\phi\left( p_{i} \right)} \right\|^{2}$ is the regular term to ensure the smoothness of the deformation field $\phi$, $\lambda$ is empirically set to 1.0.

With the constraint on the consistency of image and feature-aware content, the reg-model can better align the anatomical structures and further provide good initialization results for the subsequent image-aligned style transformation.

\subsection{Image-aligned style transformation in segmentation for better performance }


After training the reg-model, the atlas image $I_a$ is warped by the predicted deformation field $\phi$ to obtain the warped atlas ${I_{\tilde{a}}}$, which is roughly aligned to the unlabeled image $I_{u}$. Meanwhile, the label of the atlas $S_{a}$ is warped by $\phi$ accordingly, obtaining the pseudo mask $S_{a} \circ {\phi}$. For the convenience of unified representation, we denote the $S_{a} \circ {\phi}$ as $S_{\tilde{a}}$. We first transfer the style of $I_{\tilde{a}}$ to that of $I_u$ via image-aligned style transformation, and then utilize the transferred $I_{\tilde{a}}$ with its mask $S_{\tilde{a}}$ to train the seg-model.


\subsubsection{Image-aligned Style Transformation} 
Given an image $x \in \mathbb{R}^{W\times H\times D}$, where $W\times H \times D$ represents the image size. The Fourier transform of image $x$ can be formulated as:

\begin{equation}\label{FACC}
\resizebox{1.0\columnwidth}{!}{
	$\mathcal{F}(x)\left( {i,j,k} \right) = {\sum\limits_{w = 0}^{W - 1}{\sum\limits_{h = 0}^{H - 1}{\sum\limits_{d = 0}^{D - 1}{x\left( {w,h,d} \right)e^{- j2\pi{({\frac{w}{W}i + \frac{h}{H}j + \frac{d}{D}k})}}}}}}.$
 }
\end{equation} 

The inverse Fourier transform is defined as $\mathcal{F}^{- 1}(x)$ accordingly. Using $R(x)$ and $I(x)$ to represent the real and imaginary parts of $\mathcal{F}(x)$, the amplitude and phase components of the Fourier spectrum can be formulated as:

\begin{equation}\label{FACC}
\begin{aligned}
    &\mathcal{A}(x) = \sqrt{R^{2}(x) + I^{2}(x)},\\
    & {\rm and}\ \mathcal{P}(x) = \arctan \left( \frac{I(x)}{R(x)} \right).
\end{aligned}
\end{equation}

The method \cite{xu2021fourier} observed that the phase component of the Fourier spectrum retains high-level semantic information such as the spatial structure and shape of the original image, while the amplitude component mainly contains low-level statistical information such as grayscale and style. Inspired by this observation, we attempt to exchange the amplitude component of the atlas and unlabeled images to transfer the style of the atlas.

However, since the amplitude component also contains a small amount of high-frequency information, such as image edges, directly exchanging the amplitude component of two unaligned images inevitably introduces additional artifacts, which will reduce image quality and bring about the mismatching between the transformed atlas and its mask. This may influence the performance of the subsequent segmentation training. Thus, we instead first roughly align the atlas and the unlabeled image through reg-model, and then transplant the amplitude component of the unlabeled image $I_u$ into that of the warped atlas ${I_{\tilde{a}}}$ to transform the style of the warped atlas.

Figure \ref{fourier}(a) illustrates the detailed description of image-aligned style transformation (IST). After registration, we can obtain a warped atlas $I_{\tilde{a}}$ roughly aligned with the unlabeled image $I_{u}$, we then first perform Fourier transform on $I_{\tilde{a}}$ and $I_{u}$,  respectively, to obtain their corresponding components of phase and amplitude, which are denoted as $\left\{ {\mathcal{P}\left( I_{\tilde{a}} \right),\mathcal{A}\left( I_{\tilde{a}} \right)} \right\}$ and $\left\{ {\mathcal{P}\left( I_{u} \right),\mathcal{A}\left( I_{u} \right)} \right\}$. Denoting the transformed warped atlas as $\overline{I_{\tilde{a}}}$, we preserve the phase of $I_{\tilde{a}}$ as that of $\overline{I_{\tilde{a}}}$, i.e., $\mathcal{P}\left( \overline{I_{\tilde{a}}} \right) \gets \mathcal{P}\left( {I_{\tilde{a}}} \right) $, and mix-up the amplitude of $I_{\tilde{a}}$ and that of $I_{u}$ with a perturbation factor to obtain the amplitude components of $\overline{I_{\tilde{a}}}$:

\begin{equation}\label{FACC}
	\mathcal{A}\left(\overline{I_{\tilde{a}}} \right) = \left( {1 - \beta} \right) \times \mathcal{A}\left( I_{\tilde{a}} \right) +  \beta \times \mathcal{A}\left( I_{u} \right),
\end{equation} 
where $\beta$ is a random number uniformly distributed between 0 and 1.

Based on $\overline{I_{\tilde{a}}}$'s amplitude component $\mathcal{A}\left(\overline{I_{\tilde{a}}} \right)$ and phase component $\mathcal{P}\left( \overline{I_{\tilde{a}}} \right)$, the Fourier spectrum of $\overline{I_{\tilde{a}}}$ can be obtained:

\begin{equation}\label{FACC}
	\mathcal{F}\left( \overline{I_{\tilde{a}}} \right)\left( {i,j,k} \right) = \mathcal{A}\left( \overline{I_{\tilde{a}}} \right)\left( {i,j,k} \right)e^{- j \times \mathcal{P}{(\overline{I_{\tilde{a}}})}{({i,j,k})}}.
\end{equation} 

We final perform inverse Fourier transform on $\mathcal{F}\left( \overline{I_{\tilde{a}}} \right)$ to obtain the transformed warped atlas:

\begin{equation}\label{FACC}
	\overline{I_{\tilde{a}}} = \mathcal{F}^{- 1}\left( {\mathcal{F}\left( \overline{I_{\tilde{a}}} \right)} \right).
\end{equation} 

We would highlight the advantages of IST: 1) We preserve the phase of ${I_{\tilde{a}}}$ so the spatial structure information of $\overline{I_{\tilde{a}}}$ can be consistent with ${I_{\tilde{a}}}$, thus ensuring the correspondence between $\overline{I_{\tilde{a}}}$ and the pseudo masks $S_{\tilde{a}}$ ($I_{\tilde{a}}$ and $S_{\tilde{a}}$ are spatially aligned). 2) The amplitude of the unlabeled image is embedded into that of $\overline{I_{\tilde{a}}}$, so the image style of $\overline{I_{\tilde{a}}}$ can be similar to that of the unlabeled image $I_{u}$, preserving the diversity of training data.

\subsubsection{Segmentation details}
The transformed warped atlas image $\overline{I_{\tilde{a}}}$ is perfectly aligned with its label $S_{\tilde{a}}$, and through IST we can obtain a large number of aligned image-label pairs, in which the image style is similar to the unlabeled image. Therefore, we exploit these data to train the seg-model in a supervised manner, and the loss function between the prediction and the label can be calculated as: $\mathcal{L}_{Seg} = - Dice({{\hat{S}}_{\tilde{a}}, S_{\tilde{a}}})$, where ${\hat{S}}_{\tilde{a}}$ represents the predicted segmentation results of the transformed warped atlas image  $\overline{I_{\tilde{a}}}$. $Dice(\cdot,\cdot)$ is the Dice coefficient, which is a metric to measure the coincidence degree between two sets. The Dice of two masks, i.e., $A$ and $B$, can be formulated as:

\begin{equation}\label{Dice}
	Dice(A,B)=\frac{2|A\cap B|}{|A|+|B|}.
\end{equation}





With this aligned and diverse data, we leverage the powerful learning capabilities of CNNs, i.e., the segmentation network, to establish pixel-wise mappings between images and segmentation results, thus providing more accurate guidance for subsequent iterations.

\subsection{Iterative training of registration and segmentation}
After training the seg-model, given an unlabeled image, the segmentation results of the seg-model, i.e., ${\hat{S}}_{u}$, will theoretically be closer to the ground truth than that obtained from the reg-model, i.e., $S_{\tilde{a}}$. Therefore, in the subsequent training of the reg-model, the seg-model provides the reg-model with the predicted segmentation results of the unlabeled images, which are utilized as the additional auxiliary labels to weakly supervise the training process of the reg-model.

Specifically, we exploit the negative Dice coefficient between $S_{\tilde{a}}$ and ${\hat{S}}_{u}$ as the weak supervision term for training the reg-model: $\mathcal{L}_{Weak} = - Dice({S_{\tilde{a}},\hat{S}}_{u})$.


Therefore, the loss function of the reg-model for the first iteration can be formulated as:

\begin{equation}\label{FACC}
	\mathcal{L}_{Reg}^{1} = \mathcal{L}_{Reg}^{0} + \mathcal{L}_{Weak}.
\end{equation} 

The `reg-seg-reg' training process is then repeated until the network converges. In this way, through iterative training, the reg-model and seg-model mutually utilize the additional information provided by each other to gradually improve the registration and segmentation accuracy.

\subsection{Implementation Details }
We adopt the current state-of-the-art registration method \cite{lv2022joint} as our reg-model, and a 3D-UNet \cite{3dunet} with a strategy of deep supervision as our seg-model. For both the unsupervised  (initial) and weakly supervised (iterative) training phase of the reg-model, the learning rate was set to 1×$10^{-4}$, and the training was performed for 40,000 steps. We trained the seg-model for 20,000 steps with a learning rate of 1×$10^{-3}$. During training, we performed random spatial transformations including affine and B-spline transformations on each training image to enhance the robustness of the network. We implemented our method based on Tensorflow \cite{tf} and used the Adam optimizer to train the network. All training and testing were performed on a GPU resource of NVIDIA RTX 3090, and a CPU resource of Intel Xeon Gold 5220R.

\section{Experiments and Results}
\subsection{Datasets and Evaluation Metrics}
\subsubsection{OASIS} The dataset \cite{oasis} contains 414 scans of T1 brain MRI with an image size of 256×256×256 and a voxel spacing of 1×1×1 mm. The ground-truth masks for 35 brain tissues are obtained by FreeSufer \cite{freesurfer} and SAMSEG \cite{puonti2016fast}.

\subsubsection{CANDIShare} The dataset \cite{CANDI} contains 103 scans of T1 brain MRI with an image size ranging from 256×256×128 to 256×256×158, and the voxel spacing is around 1×1×1.5 mm. The dataset also provides labeled data of multiple brain structures. For a fair comparison with \cite{wang2020lt}, we referred to their paper \cite{wang2020lt} and selected 28 brain tissues for experiments.

\begin{table}\centering
	\renewcommand\arraystretch{1.3}
    \resizebox{1.0\columnwidth}{!}{
	\begin{tabular}{l|c|c} 
		\toprule
		Method & OASIS & CANDIShare \\
		\midrule
		Elastix \cite{klein2009elastix}&$0.750\pm0.019$ & $0.785\pm0.016$  \\
		VoxelMorph \cite{vm}&$0.787\pm0.019$ & $0.802\pm0.018$  \\
		RCN \cite{RCN}& $0.798\pm0.018$ & $0.805\pm0.015$  \\
		PCNet \cite{lv2022joint}& $0.808\pm0.014$ & $0.812\pm0.013$  \\
		Brainstorm \cite{zhao2019data}&$0.813\pm0.018$ & $0.827\pm0.015$  \\
		DeepAtlas \cite{xu2019deepatlas}&$0.819\pm0.018$ & $0.828\pm0.014$  \\
		LT-Net \cite{wang2020lt}&- & $0.823\pm0.025$  \\
		SiB \cite{wang2021alternative}&- & $0.830\pm0.018$ \\
		Ours & $\bm{0.851\pm0.017}$ & $\bm{0.839\pm0.012}$  \\
		\bottomrule
	\end{tabular}}
	\caption{Comparison results with traditional method and other state-of-the-arts on OASIS and CANDIShare dataset. The best performance is marked in bold.}
	\label{tab_1}
\end{table}

\begin{figure}[!t]
	\centerline{\includegraphics[width=1.0\columnwidth]{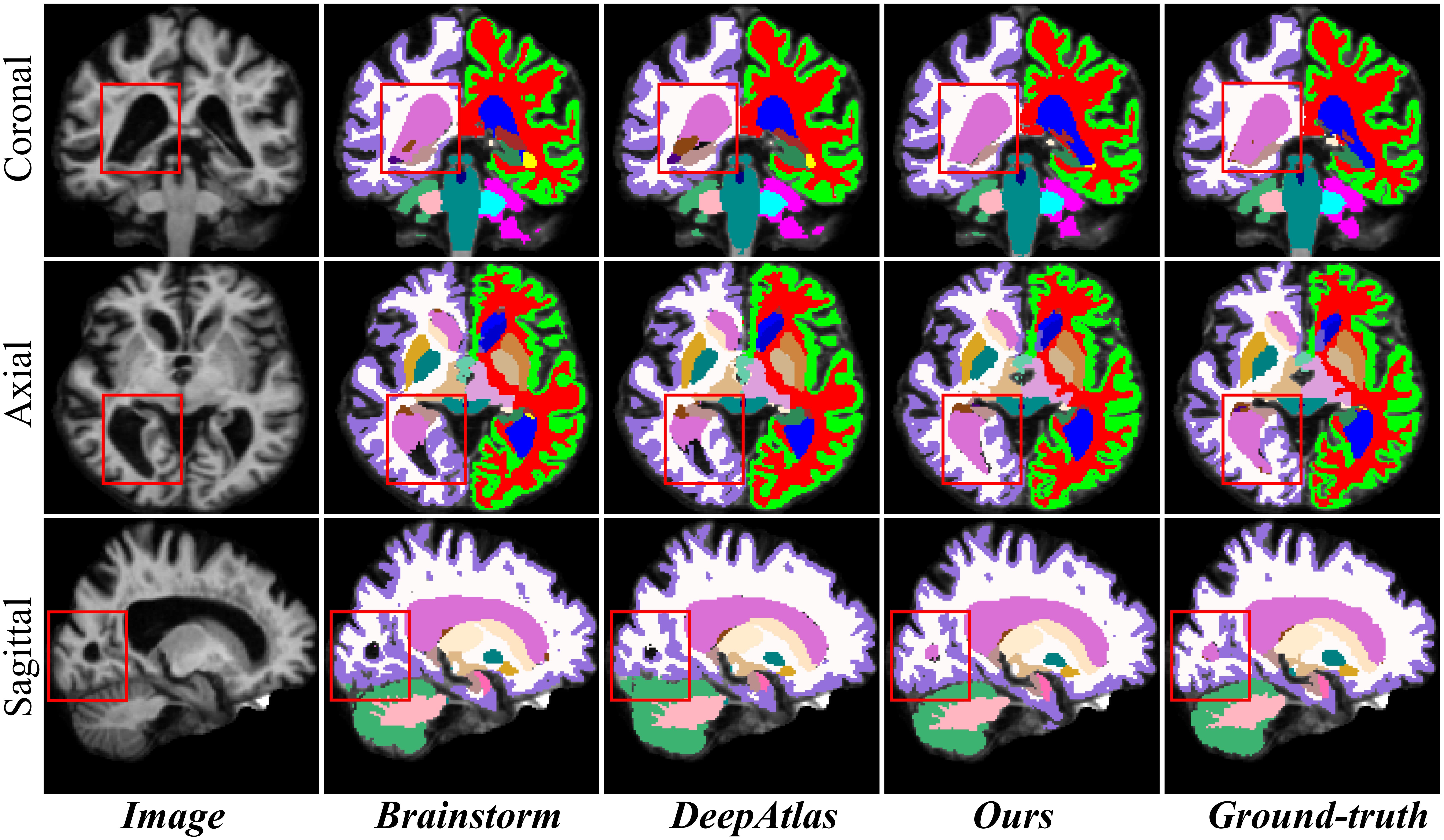}}
	\caption{
	The visualization results of different one-shot segmentation methods trained with one labeled image.
	}
	\label{visulization}
\end{figure}

All MRI images were pre-processed to remove the skull, correct the bias field, and transfer to a template space via rigid registration. We randomly divided the data in each dataset into training and test sets, obtaining 331 training and 83 test images in OASIS, and 83 training and 20 test images in CANDIShare. For each of OASIS and CANDIShare, according to \cite{vm} and \cite{wang2020lt}, we chose the atlas from the training data which has the highest image-level similarity with the test set. The similarity is calculated by first computing the NCC score between the atlas and each test image, and then averaging all scores. The rest of the training set was regarded as unlabeled images. Due to the limitation of GPU memory, the images from both two datasets were all re-sampled to the size of 128×128×128.


\subsubsection{Evaluation Metrics} We evaluated the registration and segmentation methods with the Dice coefficient, which is defined in Eq.(\ref{Dice}). The range of Dice is from 0 to 1, and when two images are registered perfectly, the Dice is 1.



\subsection{Comparison with the state-of-the-arts}
We compared our method with one traditional method, i.e., Elastix \cite{klein2009elastix}, and 7 state-of-the-art (SOTA) methods utilized for one-shot segmentation, i.e., VoxelMorph \cite{vm}, RCN \cite{RCN}, PCNet \cite{lv2022joint}, Brainstorm \cite{zhao2019data}, DeepAtlas \cite{xu2019deepatlas}, LT-Net \cite{wang2020lt} and SiB \cite{wang2021alternative}. Among them, Brainstorm and DeepAtlas are dual-model iterative methods, and others are registration methods. Table \ref{tab_1} shows the average Dice achieved by our method and other comparison methods on the OASIS and CANDIShare datasets. The Elastix, VoxelMorph, RCN, PCNet, Brainstorm, and DeepAtlas were implemented through their released source code. LT-Net and SiB have no source code provided, we borrowed the results from their papers, which only provided results on the CANDIShare dataset. It is worth mentioning that the CANDIShare dataset was divided in the same way as described in LT-Net \cite{wang2020lt} and SiB \cite{wang2021alternative}.

It can be seen from Table \ref{tab_1} that our method outperforms the SOTA methods on both datasets. Compared with the SOTA dual-model iterative methods, our method improves the average Dice by at most 4.67\% and 1.45\% on OASIS and CANDIShare, respectively. It indicates that our method can better overcome the problem of lacking labeled data, and can achieve better segmentation results than the current state-of-the-arts. We also visualize the segmentation results of different comparison methods on the OASIS dataset in Figure \ref{visulization}. We can observe that the segmentation results of our method are closest to the Ground-truth in some small regions, which demonstrates the effectiveness of our method.


\subsection{Ablation Studies}

\subsubsection{Effectiveness of Feature-aware Content Consistency} 

\begin{figure}[t]
	\centerline{\includegraphics[width=1.0\columnwidth]{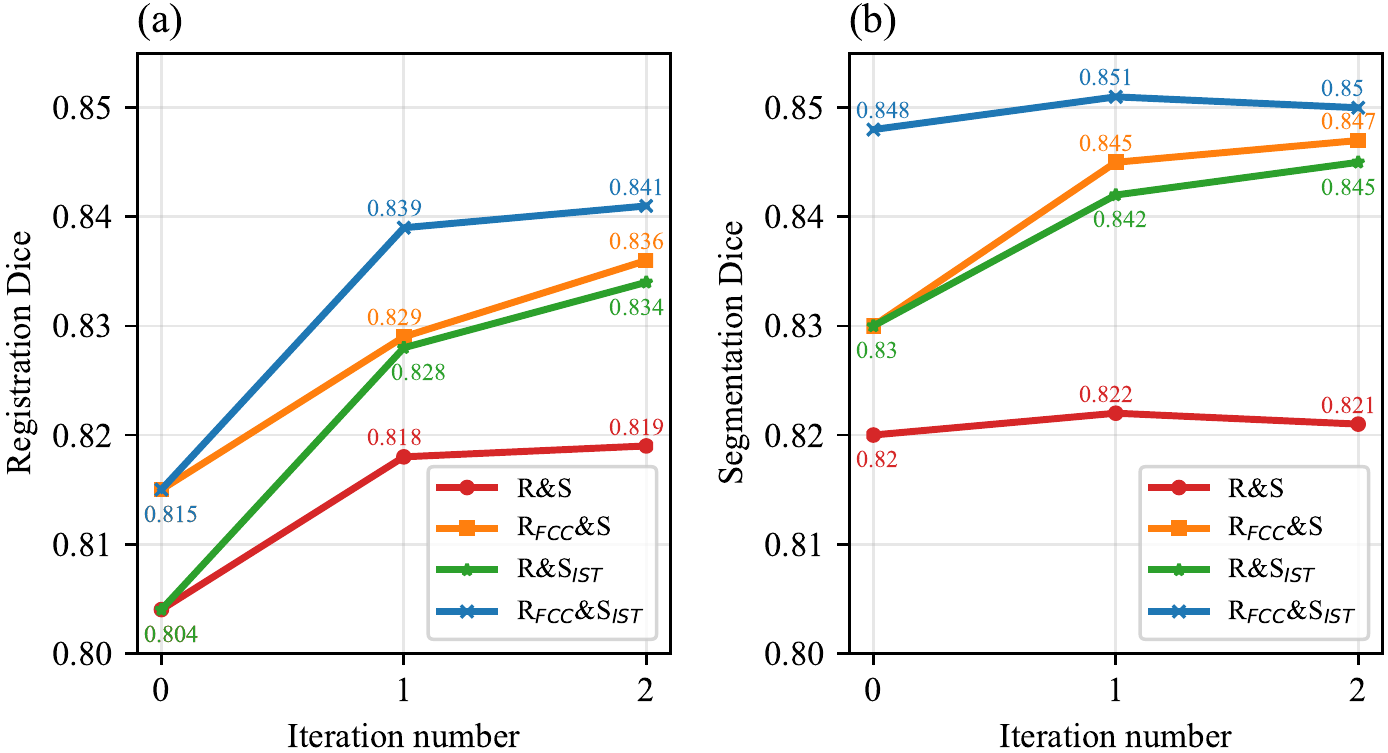}}
	\caption{The average Dice for (a) registration and (b) segmentation of different variants in ablation studies. The R\&S,  R$_{FCC}$\&S, R\&S$_{IST}$, and R$_{FCC}$\&S$_{IST}$ represents variant of pure reg-model \& pure seg-model, reg-model with FCC \& pure seg-model, pure reg-model \& seg-model with IST, and reg-model with FCC \& seg-model with IST, respectively.}
	\label{iter_fig}
\end{figure}

To verify the effectiveness of the feature-aware content consistency,  we conduct an ablation experiment by adding/removing feature-aware content consistency (FCC) in the reg-model. The results of R\&S and R$_{FCC}$\&S in Figure \ref{iter_fig} show the effect of feature-aware content consistency in the performance of registration and segmentation. 

Comparing the registration performance of R\&S and R$_{FCC}$\&S (Figure \ref{iter_fig}(a)), we can observe that after 3 iterations, the reg-model trained by optimizing the combination of image consistency and feature-aware content consistency achieves better results than that only trained by image consistency, with an improvement of average Dice by 1.81\% (p-value \textless 0.001). Besides, the segmentation performance (Figure \ref{iter_fig}(b)) of R$_{FCC}$\&S is at most 2.69\% (p-value \textless 0.001) higher than that of R\&S. These results show that additionally constraining the feature-aware content consistency can effectively promote the registration and segmentation performance of our iterative dual-model.

\subsubsection{Effectiveness of Image-aligned Style Transformation} 



To further verify the effectiveness of image-aligned style transformation, we also perform a comparison experiment in which we trained the seg-model with or without image-aligned style transformation, that is, we trained two seg-models based on different training data, one is trained with the misaligned pairs of unlabeled image $I_u$ and pseudo mask $S_{\tilde{a}}$, and the other is trained with the aligned pairs of transformed warped atlas $\overline{I_{\tilde{a}}}$ and its label $S_{\tilde{a}}$.

Comparing R\&S with R\&S$_{IST}$ in Figure \ref{iter_fig}, it can be observed that after three iterations, the final segmentation performance of the seg-model trained with the aligned image-label pairs is better than that trained with the misaligned pairs, with an improvement of average Dice by 2.51\% (p-value \textless 0.001). It indicates that the spatial misalignment in the image-label pair negatively affects the seg-model, resulting in an inferior segmentation performance. In contrast, the image transformed via IST not only guarantee the pattern diversity of the unlabeled image but also avoids the spatial misalignment with the label, thus, not surprisingly, the accuracy of segmentation can be significantly improved. This significant improvement confirms the important roles of spatial aligned image-label pair in one-shot segmentation and demonstrates the effectiveness of our proposed image-aligned style transformation.


\begin{figure}[t]
	\centerline{\includegraphics[width=0.77\columnwidth]{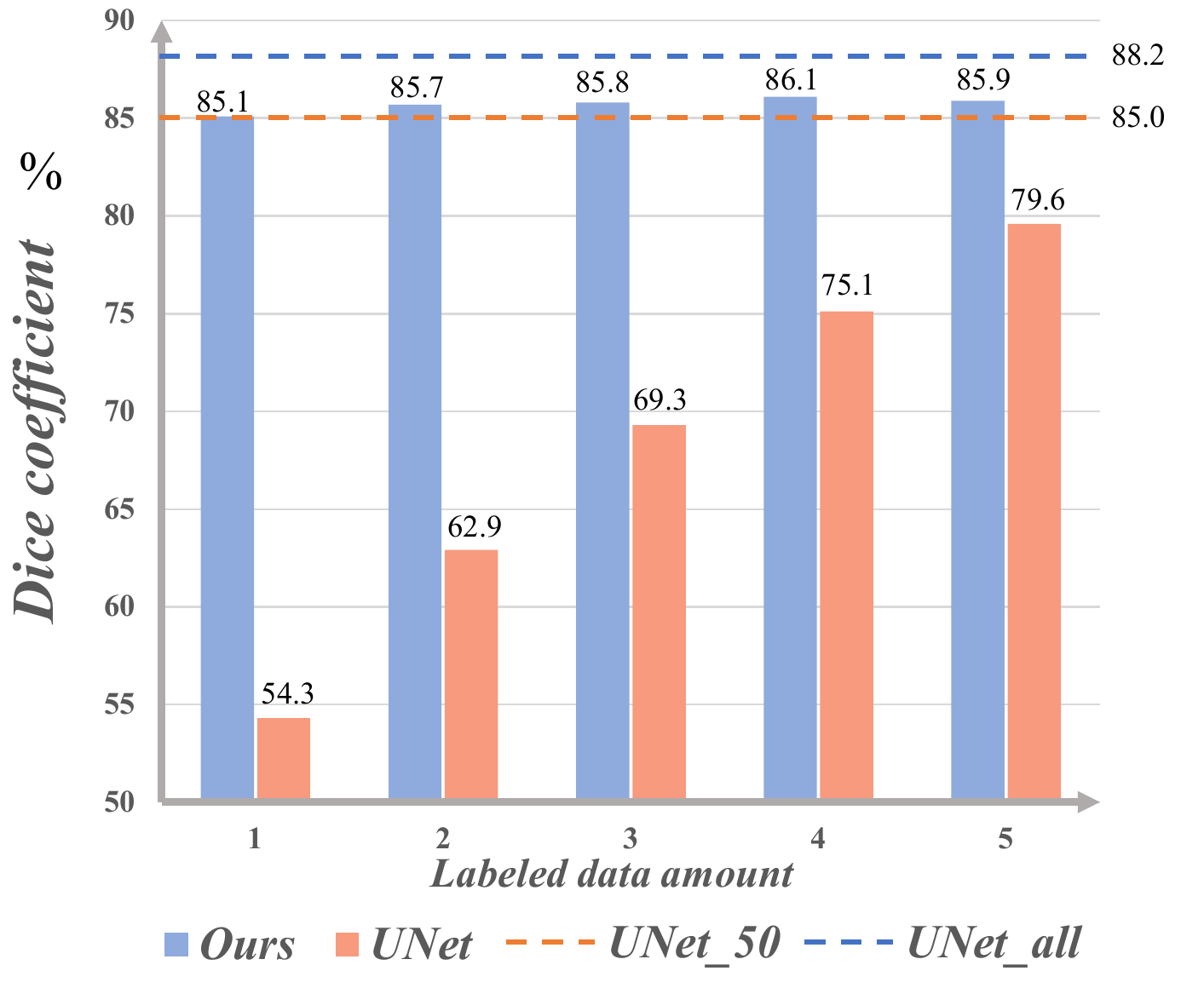}}
	\caption{Comparison results of the segmentation accuracy achieved by our method and that by a vanilla segmentation network trained with different numbers of labeled atlases. UNet\_all represents the performance of 3D-UNet trained with all labeled training images (331) on the OASIS dataset.}
	\label{seg_results}
\end{figure}



\subsubsection{The effect of labeled data amount on the segmentation results} 

Based on the OASIS dataset, we compared our method and a vanilla segmentation network (3D-UNet). We trained the two methods several times and each time used different numbers of labeled atlases to train. Figure \ref{seg_results} illustrates the change of the average Dice when the number of labeled samples increases from 1 to 5.


It can be seen from Figure \ref{seg_results} that the segmentation accuracy of our method is higher than that of 3D-UNet with the same number of available labeled atlases. By comparing Ours (n=5) and UNet\_5 (3D-UNet trained with 5 labeled images), we can observe that our method improves the average Dice by 7.91\%. With only one labeled atlas, our method achieves comparable segmentation accuracy to the supervised 3D-UNet using 50 annotated samples. The promising result indicates the high potential of our method in the absence of annotation data.


\begin{figure}[!t]
	\centerline{\includegraphics[width=0.91\columnwidth]{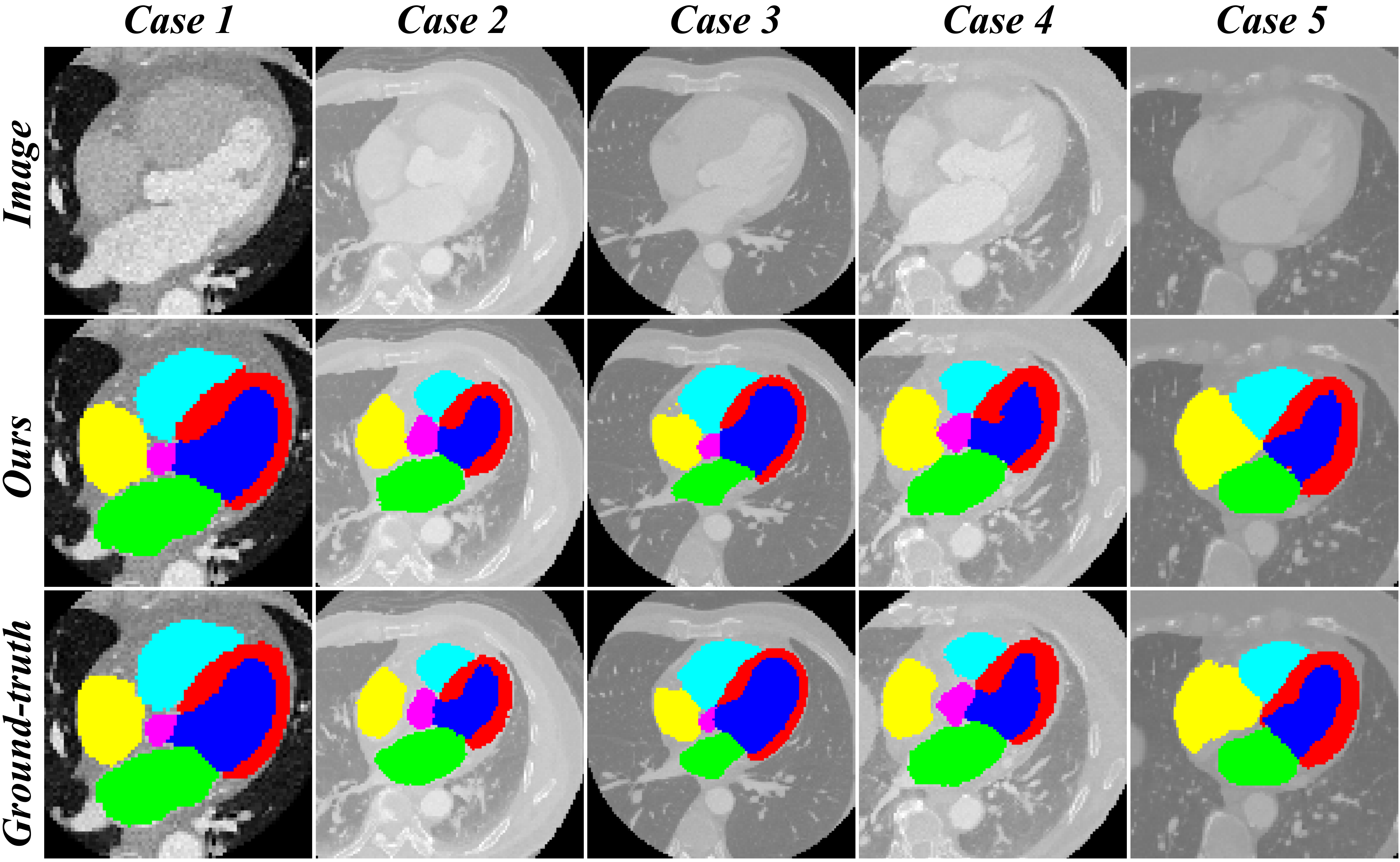}}
	\caption{
	The visualization results of our method on the CT heart dataset \cite{zhuang2016multi}.}
	\label{visulization_ct}
\end{figure}

\subsection{Scalability to Other Modalities}
We also applied our method to other modalities, i.e., 3D heart CT MH-WH 2017 dataset \cite{zhuang2016multi} to evaluate the scalability of our method, that is, the employed seg/reg models are not technically restricted to a specific modality/organ as long as registration can be performed. We randomly selected one labeled image (atlas) together with the 40 unlabeled images as the training set, and the remaining 19 labeled images were utilized as the test set. Our method achieves $0.886 \pm 0.024$ Dice on the test set. We also randomly selected the segmentation results of five cases from the test set, and the visualization results are shown in Figure \ref{visulization_ct}. It can be seen that our method also shows good performance on the CT images of the human heart, verifying our promising scalability to other modalities.

\section{Conclusion}

In this paper, we propose an image-aligned style transformation to address the degradation of one-shot segmentation performance caused by the spatial misalignment of the image-mask pair. We also introduce a feature-aware content consistency to further avoid the collapse of image-aligned style transformation in the first iteration. Experiments on two public datasets, OASIS and CANDIShare, demonstrate the effectiveness of our developed one-shot segmentation framework. The ablation studies also prove the effectiveness of our proposed two strategies, i.e., image-aligned style transformation and feature-aware content consistency. With only one labeled image, our method can achieve comparable accuracy to the segmentation network trained on 50 labeled images, which shows the great potential application in the situation of high labeling costs. The promising results on the 3D CT MH-WH 2017 dataset demonstrate the good scalability of our method to other modalities. Our further work will focus on the more challenging task, i.e., the one-shot abdominal multi-organ segmentation.

\clearpage
\section{Acknowledgments}
This work was supported in part by National Key R\&D Program of China (Grant No. 2022YFE0200600), National Natural Science Foundation of China (Grant No. 62202189), Fundamental Research Funds for the Central Universities (2021XXJS033), Science Fund for Creative Research Group of China (Grant No. 61721092), Director Fund of WNLO, Research grants from United Imaging Healthcare Inc.

\bibliography{aaai23}
\end{document}